# In-the-wild Facial Expression Recognition in Extreme Poses

Fei Yang[1], Qian Zhang[2], Chi Zheng[3] and Guoping Qiu[4]

International Doctoral Innovation Centre, University of Nottingham Ningbo China[1]
School of Computer Science, University of Nottingham Ningbo China[2]
Institute of Microscopy Science and Technology, Ningbo Yongxin Optics Co. LTD.[3]
School of Computer Science, University of Nottingham[4]

## ABSTRACT

In the computer research area, facial expression recognition is a hot research problem. Recent years, the research has moved from the lab environment to in-the-wild circumstances. It is challenging, especially under extreme poses. But current expression detection systems are trying to avoid the pose effects and gain the general applicable ability. In this work, we solve the problem in the opposite approach. We consider the head poses and detect the expressions within special head poses. Our work includes two parts: detect the head pose and group it into one pre-defined head pose class; do facial expression recognize within each pose class. Our experiments show that the recognition results with pose class grouping are much better than that of direct recognition without considering poses. We combine the hand-crafted features, SIFT, LBP and geometric feature, with deep learning feature as the representation of the expressions. The handcrafted features are added into the deep learning framework along with the high level deep learning features. As a comparison, we implement SVM and random forest to as the prediction models. To train and test our methodology, we labeled the face dataset with 6 basic expressions.

## 1. INTRODUCTION AND REALTED WORK

With the development of human computer interaction, facial expression is one of the most important method for machine to understand human behaviors. Face expression contains much information of the participants and the audience will obtain much "words" from that. To understand the meaning of expressions is a great gap and difficult for computers, which is an important part of human behavior modeling. Facial expression recognition has been a hot research topic for a long time and much work has been done. Automatically facial expression recognition is a challenging task. We can find much work in literature talking about this topic. But most of them is mainly about the frontal faces and the expression is captured from volunteers under controlled condition, not in-the-wild. Recently, as the recognition for frontal face has shown better and better performance, much attention is focused on the expression recognition on different poses or under different circumstances. In our work, we use in-the-wild data set for both training and testing set.

There are two main approaches to recognize face expressions. One is to extract the features, texture features or geometric features, which are related to face expressions and then use the classifiers to train the classification models. The other is to use the Facial Action Coding System(FACS) defined by Ekman and Friesen in 1977[1][2] to make for a description of face local regions. The combination of action units is set as the representation of expressions. But no matter what kind of approach we take, the concept of the process is the same: get the representation of the expressions and then train the models to recognize and classify.

Researchers normally classify facial expression into six categories: happiness, sadness, anger, fear, surprise, and disgust. The reader can refer to [3][4] for other types of facial expressions. S Du and Y Tao collected a data set from 230 participants to have a series of expressions for 21 categories [5]. In future, more research work should be done on multiple or compound facial expressions. But at present, many researchers are still trying to solve the problems for the 6 basic ones with the issues under different complex conditions, such as light, head pose, fake expression, race, and etc.

Traditionally the process to recognize an expression usually consists of the following steps: extract features from the photo, pass them to the classifier, and get the result. The main problems would be how to get a good feature for expression representation and how to train a good classifier to classify them. There are usually two features for the face expression representation, the texture feature and the geometric feature. Both have shown their great power in facial expression representation. In our work, we combine these two different features to achieve a better performance. SVM and random forest are both very good classifiers. They have different classification theories and also have different strengths and drawbacks. We tested both with different features and in the final system, the results from different classifiers are combined to give a better decision.

Using the features extracted from faces as the representation of facial expressions is a common way [6]. For decades of research, many kinds of features have been derived, such as HOG, SIFT, SURF, Wavelet, Gabor, DCT and etc. For different recognition problems, they may have different strengths. Specially, The Local Binary Pattern (LBP) feature has been much used in face recognition problem [7][8]. It shows great power in overcoming the illumination effects and costs less computation, which is more applicable in real time system. Some applications can be referred [9][10]. The combination of texture feature and geometric feature is also used [11][12][13] and it was approved to achieved a better performance than only using a single feature.

Convolutional neural network has been proved to perform very well in a number of research areas, such as object detection [14][15][16][17], face recognition[18][19][20], and speech recognition[21][22]. In current computer vision research, we can seldom avoid talking about CNN. Deep learning feature has much prior ability than traditional hand-crafted features. By training with large amount of data, the network is able to represent the expressions in a high-level representation well. In our work, we use Caffe to build the network and set two phases to train the network with hand-crated features input. The results are compared with that of SVM and random forest. The hand-crafted features we talked above will be sent to the network to training together with the network. There is a philosophy we consider here. Though we have design the network as small as possible – smaller net will have less ability to grasp feature and simulate the non-linear transformation, our data may be still not enough to train the net well. The hand-crafted feature has been proved to be effective and we believe that the current network training will grasp different features with the hand-crafted feature. The combination of the hand-crafted feature should give the network additional effective information to make better decision.

In this work, face detection and face alignment is not our point. We would not talk too much about how to detect the face and how to do the preprocessing. We suppose the face is well detected and cropped from the original image. This paper mainly includes the following parts: Firstly, we talk about the process of face landmark; Secondly, we talk about the features that are combine into our proposed deep learning framework; Thirdly, we introduce our proposed CNN framework; and finally, we present the experiment results. The aim of this work is to seek the possible effective path to do facial expression recognition in-the-wild where different poses are available.

Our contribution can be summarized:

- we propose a new view to deal with in-the-wild facial expression recognition. Instead of avoiding and overcoming the head pose impacts, we do in the opposite approach to detect the head pose and do the expression recognition under pose awareness.
- we labeled the face dataset with expression labels.
- we proposed a convolutional network to predict expression with head pose oriented condition. We explore the way to improve the detection accuracy under extreme pose keeping low level computation.
- Traditional features are tested and our experiments shows the combination of hand-crafted features and the deep learning feature can improve the expression representation performance.

## 2. HEAD POSE DETECTION AND CLASS GROUPING

### 2.1 Face landmark

Face alignment is an important step of face processing problem. The common approach is to detect the landmark of one face. The definition of face landmark is to locate the key points of one face, indicating the location of eyes, mouth, nose and so on, as Figure 1. There are many different definitions of the key points, such as the types of 5 points, 15 points, 30 points, 54 points, and 68 points. In this work, we use the 68 key points face landmark.

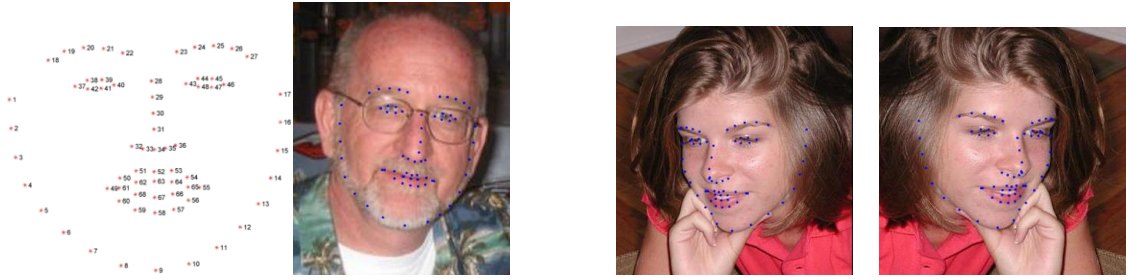

Figure 1. 68 points face landmark and Photo flipping examples. The landmark points of the flipping photo change the order. So the points need to be reordered.

We use ASM approach to detect the key points. The landmark is used for two purposes. One is to predict the head pose. The other is to generate features, the geometric feature and the SIFT feature around the key points.

## 2.2 Procrustes

Obviously, head pose is one factor that causes the difference in landmark. We take this property which could be used for clustering head poses. But except for the head pose, there are many other factors that also affect the difference of landmarks, such as face position in the image, scale, 2D rotation, expression, identification. We believe that the first three, translation, scale, and 2D rotation, may have relatively bigger effects. Procrustes process can be a solution to eliminate these three effects, which has been talked in many works [23][24][25] [26].

One landmark can be represented as $(x_1, y_1), (x_2, y_2) \ldots (x_{68}, y_{68})$, in the training set, we can calculate the mean landmark $(\overline{x_1}, \overline{y_1}), (\overline{x_2}, \overline{y_2}) \ldots (\overline{x_{68}}, \overline{y_{68}})$

The Procrustes
1) Translation: Calculate the center of the landmark and make the center zero position.

Calculate the mean of the landmark $\overline{x} = \frac{1}{68}(x_1 + x_2 + \ldots + x_{68}), \overline{y} = \frac{1}{68}(y_1 + y_2 + \ldots + y_{68})$. To translate the center to zero point, we get the new landmark for each point: $x_{new} = x - \overline{x}, y_{new} = y - \overline{y}$

2) Scaling:

The scale of the landmark is defined: $s = \sqrt{\frac{(x_1 - \overline{x})^2 + (y_1 - \overline{y})^2 + \ldots + (x_{68} - \overline{x})^2 + (y_{68} - \overline{y})^2}{68}}$

Compare each landmark with the mean landmark to have the scale value $s_{ratio} = s / s_{mean\_landmark}$, and then rescale the landmark points $((x_i - \overline{x}) / s_{ratio}, (y_i - \overline{y}) / s_{ratio})$

3) Rotation:
After remove the effects of translation and scaling, if rotate the image by θ, we can get the rotated points
$(u_1, v_1) = (\cos\theta \cdot x_1 + \sin\theta \cdot y_1, \sin\theta \cdot x_1 + \cos\theta \cdot y_1)$

The difference between two landmarks can be expressed in the sum of squared distances:
$d = (u_1 - x_{c1})^2 + (v_1 - y_{c1})^2 + \ldots + (u_{68} - x_{c68})^2 + (v_{68} - y_{c68})^2$ . $(x_{ck}, y_{ck})$ is the comparing landmark, the mean landmark.

Taking the derivative of it, the best θ can be set: $\theta = \tan^{-1}\left(\dfrac{\sum_{i=1}^{68}(x_i y_{ci} - y_i x_{ci})}{\sum_{i=1}^{68}(x_i x_{ci} - y_i y_{ci})}\right)$

In this way, the coordinate $(u_k, v_k)$ is calculated for the new landmark.

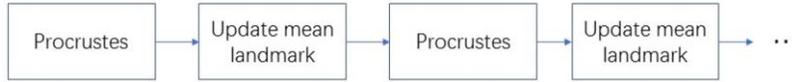

Figure 2. Iteration to do Procrustes

The Procrustes process adjusts each sample to the mean landmark. While the mean landmark changes after each Procrustes process. Therefore, multi-times iterations of Procrustes process are done with updating the mean landmark after each iteration, as Figure 2. Empirically, we set the iteration 100 times and get a relatively stable normalized landmark finally. Figure 3 shows the normalized landmark after Procrustes.

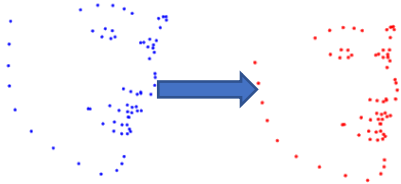

Figure 3. Landmark normalized by Procrustes process.

## 2.3 PCA projection and central landmark rechecking

After doing the Procrustes process, we think head pose is the main cause to the difference of shapes. Principal Component Analysis (PCA) is implemented [27]. We transform the landmark 68x2 into a vector and generate the landmark matrix by putting the vector of each face sample into a column. All landmarks are projected on the first eigen vector. Simply, we can set a threshold to the projected values to separate all samples into different poses.

To make the pose class classification more accurately, we calculate the central landmark for each pose class and calculate the distances for each sample to these central landmarks. And then assign the pose class by choosing the nearest central landmark. This is a process similar with k-means. Central landmarks are shown in Figure 4. The PCA projection process is like a process to choose the pre-setting centers for k-means clustering. But here we do not do the k-means clustering iteration many times. Because our purpose of the PCA projection is to fix the centers for each pose class, not to give the pre-setting centers for k-means clustering.

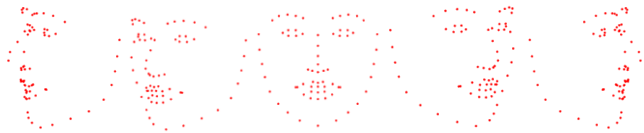

Figure 4. Five central landmarks for five pose classes.

Given a test sample, the detected face landmark is processed with Procrustes. We calculated the distance of the normalized landmark with each central landmark to decide which pose class it belongs to.

## 2.4 CNN pose classification

In fact, we cannot be very sure about the detection of the face landmark in real scenarios. Thus, besides the face landmark approach, we implement the CNN classification approach to predict the head pose. We prepare a training set to train a CNN net to predict the head pose class. The network can be designed very simple so that the computation is in a low level. The net structure is in the STRUCTURE part. In the finally system, the pose estimation is based on the two results.

# 3. HAND-CRAFTED FEATURE

## 3.1 SIFT feature

Scale-Invariant Feature Transform (SIFT) is a classical, useful and stable algorithm to find the local features in an image to describe the image. It mainly has two phases: find the key points; describe the features for the key points. The process of finding SIFT key points might be the key and brilliant part of SIFT. But here we are not doing the image matching, image retrieval or some other work which needs to find the special features of a certain image or object to express it and use these special features to do recognition problems. We only use the SIFT descriptor to extract the features for expressions representation, shown Figure 5. We do not find the SIFT points in the image. Instead, we take the 68 landmark points as the SIFT key points and extract the SIFT descriptor features around the landmark points. Another note is that we skip the step of the main direction calculation and directly get the descriptor.

For each face sample, concatenate all the SIFT feature of 68 points, as shown in figure 3.11 into a vector of 128*68=8704 long. With 122,450 samples, we have a feature matrix in the size of 122,450 by 8704. But in fact, the samples are separated into different pose classes. The training model is done within one pose class and the data set for one pose class is much smaller, about 1/5 in the case of five pose classes. In our work, pose class 1 (extreme turning right) has a sample number of 28,056, 22.9% of the data set. 19,998 of that are used for training, while this is still a large data set and it will take hours to train the SVM model on an i7-4790 CPU. Large data set is good for training as SVM will work better on larger data set. One way to reduce the computation can be to use PCA to reduce the feature dimensions. We can use PCA talked above to project all the feature into another space which can be much easier to expression the difference of samples. In the new projected space, select the first 95% contribution features to use for our training model, while in this way, the new feature dimension can be reduced to about 3,000. If select only 90%, the feature dimension can be reduced much smaller. And in this way, the performance for later SVM training doesn't change much. The transfer matrix to project the original feature with 8,704 dimension to 3,000 dimension is calculated by the selected eigen-vectors during PCA process. Given a test sample, we project it onto the same dimension space with the same transfer matrix before sending it into any classifier.

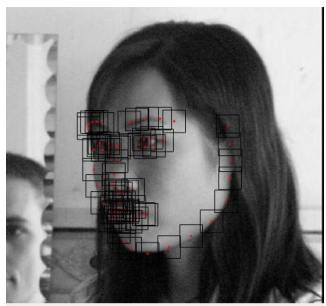

Figure 5. SIFT feature block around landmark points.

To do training better, the preprocessing of the data set is useful and necessary. Great scale changes between different dimensions is not good. We normalized the feature vector divided by the norm of the vector with zero mean.

## 3.2 Region and grid TPLBP

Local Binary Patterns (LBP) uses a 3x3 neighborhood LBP descriptor. Novel patch based LBPs, Three-Patch LBP codes (TPLBP) and Four-Patch LBP codes (FPLBP), are explained in Lior work [28]. LBP code is shown Figure 6.

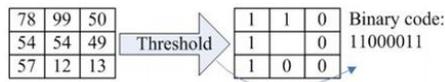

Figure 6. Basic Local Binary Pattern code. For the central pixel as comparison, compare the neighbor 8 points gray value coding with 1 if it is larger than the central pixel, otherwise 0. The 8-binary code is as the new value of the central point.

We extract TPLBP feature as the representation feature. Based on the basic LBP, it compares the central point LBP with 8 neighbors LBP feature and make a new code from the comparing. The detail derivation of TPLBP can be seen in

Lior paper [28]. We crop the face from the data set and extract the TPLBP features. We implement grid LBP. The face is separated into a 4-by-4 block and TPLBP histogram is calculated for each block. The final feature vector can be got by concatenating the histogram vectors of all blocks. Similar to grid LBP, local region LBP extracts LBP feature on special pre-defined regions. The final feature is the vector combination of all region features, as Figure 7.

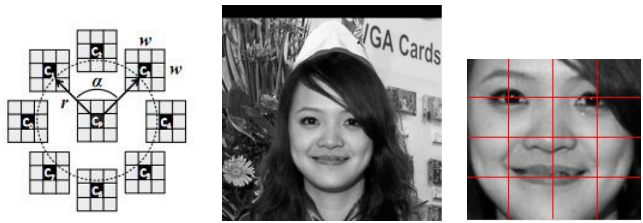

Figure 7. Three-Patch LBP and grid TPLBP blocks

### 3.3 Geometric Feature

Geometric feature is relating to the shape of the face expressed by the landmark of the face, which means we directly use the 68 points landmark information as the representation of the expressions, instead of extracting the texture feature. Each landmark contains 68 points.

We form the coordinates number into a vector as $(x_1, x_2, ..., x_{68}, y_1, y_2, y_{68})$ in the length of 136. We get the feature matrix $122{,}450 \times 136$, which is separated into different pose classes and training/test sets. The feature vector should also be uniformed as the data preparation before training.

## 4. STRUCTURE AND TRAINING

### 4.1 CNN structure

The network has three parts. The first part is the head pose prediction network. The second part is the main branch to predict the expressions. The third part is to extract the hand-crafted feature and combine these features with the network. The head pose aware network is to extract the feature which can grasp head pose information. It is different of the pose classification we talked above. The landmark head pose separation can only separate the head pose in large pose variations. But in the convolutional network, we make the output of head pose in small head pose variations, which can be implemented for each big head pose class. Our purpose is not to predict the head pose very accurately, but to guide the training to make the conv2_1 and pool2_1 to learn the head pose information. We add the low-level feature together with expression prediction network. Conv2_1 and pool2_1 are on the branch of head pose prediction net, but they are designed to base on conv1, the same with conv2_2. The low-level feature for the two branches are in the similar training direction and by using the same conv1, we can save the computation. The heatmaps after pool2 and conv2_2 are in the same size so that they can be put together. To meet this, conv2_1 is designed to keep heatmap size not changed with pad and stride of 1. Conv2_2 is designed to half size the heatmap with stride of 2. The combination of the two branches can provide more information for the final prediction. Conv3, conv4 and conv5 are layers which transfer the feature in different feature dimensions and different feature spaces. The non-linear ability makes them be able to extract good deep features. The functionality of this part is similar to the classical CNN nets like, AlexNet or CaffeNet. We reshape pool5 into one vector so that it is easy to combine with hand-crafted features. We build the net in Caffe. The framework is shown Figure 8.

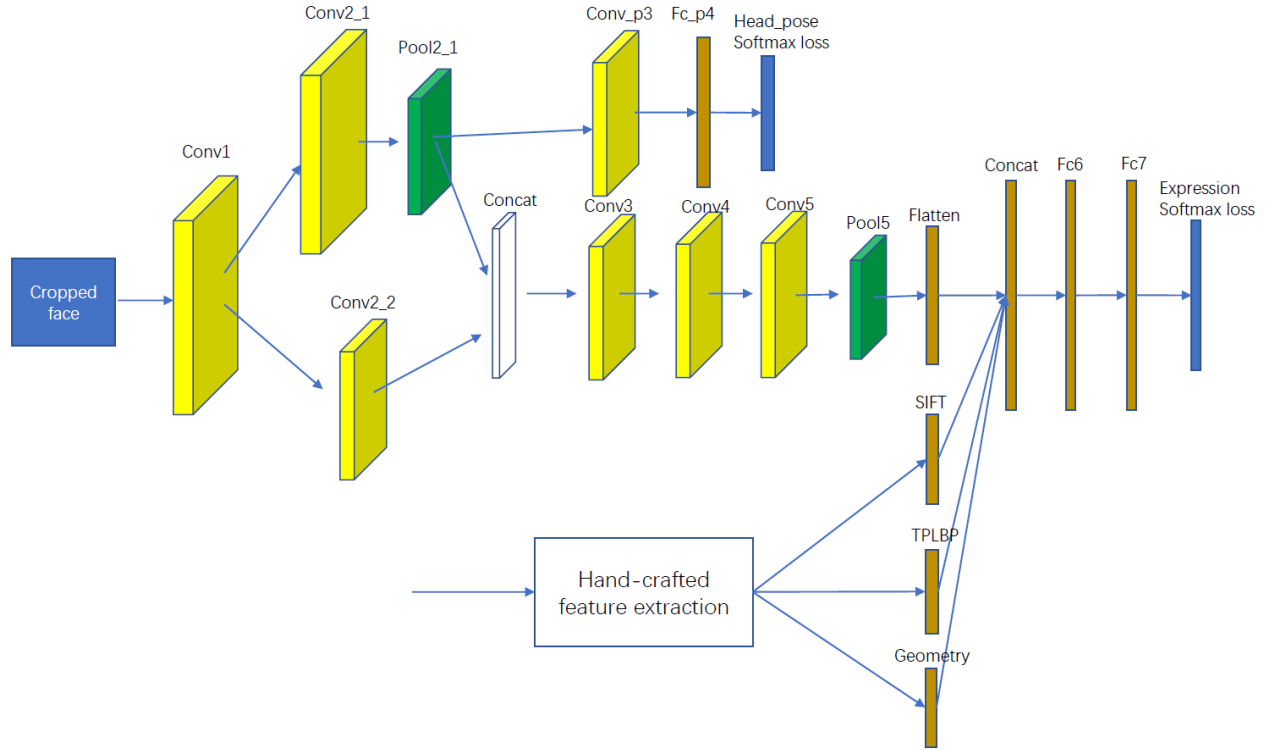

Figure 8. CNN structure to predict facial expression. The net has a branch to predict head pose. Hand-crafted feature are sent into the network. The final prediction is based on the combination of deep net feature and the hand-crafted feature.

### 4.2 Training

The hand-crafted feature is extracted in other ways, which should be processed before the network training. Therefore, the SIFT, TPLBP and geometry features are regarded as prior information sent into the network. The network is trained with two outputs, the head pose and expression classification. The first branch is supposed to gain head pose information and the main branch is to grasp the expression deep feature. While in testing phase, the image is sent to go through the first branch to predict the head pose.

## 5. EXPERIMENTS

### 5.1 Dataset

300W-LP: The 300W across Large Poses(300W-LP) database is generated from 300W data set [29] by rotating 3D head pose, created by Zhu X, Lei Z, Liu X [30][31]. 300 faces in-the-wild (300W) database consists of multiple alignment databases with 68 landmarks, including AFW [32], LFPW [33], HELEN [34], IBUG [29] and XM2VTS [35]. A general brief view is as below.

AFW: The Annotated Faces in-the-wild(AFW) database contain 250 images with 468 faces. Six facial landmark points are set for each photo.

LFPW: The Labeled Face Parts in-the-wild(LFPW) database contains 1.287 images downloaded from google.com, fickr.com, and yahoo.com. The images contain large variations including pose, expression, illumination and occlusion. For each photo, 35 landmark points have been annotated.

HELEN: The Helen database has 2.330 annotated images downloaded from Flickr. It provides much more detail annotated landmark information.

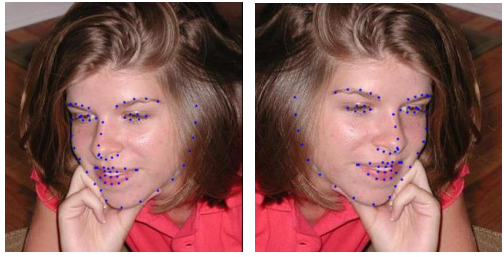

Figure 9. Photo flipping examples. The landmark points of the flipping photo change the order. The points need to be reordered.

With 300W, adopt the proposed face profiling to generate 61,225 samples across large poses (1,786 from IBUG, 5,207 from AFW, 16,556 from LFPW and 37,676 from HELEN, XM2VTS is not used), which is further expanded to 122,450 samples with flipping. The final database is as the 300W across Large Poses (300W-LP) [31].

We use the 300W-LP database with total 122,450 photos for experiment. While 70% of the images were used for training, the remaining images were used for testing. One should note that the same people faces that are generated from the same original images can only be in either training set or test set. To label the expressions for the data set, we only check the original photos, while other synthesized photos will have the same expression labels with their original ones. The data set contains the landmarks for 61,225 photos, we can also calculate the flipping photos landmark by flipping the x coordinates for the landmarks. But when we generated the landmarks for flipping photos. The order for the 68 points changed

All the photos are labeled into 7 categories, 6 basic expressions and the neutral class. The data set contains much more neutral and happy samples and fewer other expression samples, as shown in figure 3.4. Unbalanced data is not good for classification training. So better data sets can be proposed. The labeled expression numbers are shown Table 1.

| SET | Neutral | Happy | Sad | Fear | Angry | Surprise | Disgust |
|---|---|---|---|---|---|---|---|
| AFW | 52.23% | 42.14% | 0.59% | 1.19% | 0.89% | 2.67% | 0.3% |
| HELEN | 44.51% | 42.02% | 1.8% | 1.8% | 1.89% | 4.42% | 3.56% |
| IBUG | 47.41% | 28.89% | 2.96% | 0.74% | 9.63% | 8.15% | 2.22% |
| LFPW | 45.22% | 35.46% | 3.19% | 4.06% | 6.76% | 2.61% | 2.71% |
| Total | 45.48% | 39.8% | 2.11% | 2.32% | 3.39% | 3.91% | 3.0% |

Table 1. Expression labeled in data set.

## 5.2 pose class separation

The samples are separated into 5 pose classes, in which pose 3 is the set of frontal faces and pose 1,2 for turning right and pose 4,5 for turning left. During the experiment, because all the photos have flipping photos, which means pose 1,2 are much similar to pose 4,5 in a mirror flipping, it will be enough for us to just implement the experiment on pose 1,2,3. All the samples are randomly separated into training and test sets, shown Table 2.

| | Pose 1 | Pose 2 | Pose 3 | Pose 4 | Pose 5 |
|---|---|---|---|---|---|
| Train set | 19,998 | 17,934 | 11,351 | 17,934 | 19,998 |
| Test set | 8,058 | 7,206 | 4,707 | 7,206 | 8,058 |
| Total | 28,056 | 25,140 | 16,058 | 25,140 | 28,056 |

Table 2. Pose and dataset separation.

## 5.3 Some training tricks

5.3.1 Net pre-trained

CNN network training relies on huge amount of data. Though our dataset consists of 100 thousand of images. But most of them are generated from the original images. Among the 3,500 original images, most of the faces are neutral and happy. Other expression samples are in small number. Therefore, these samples are not enough to train.

The philosophy of transfer learning is implemented. We pre-train the net on other large face datasets. CASIA-Webface is used to train. It consists of 494,414 faces from 10,575 subjects. The pre-trained net is supposed to grasp the face identity features.

5.3.2 Sample balancing.

The classes of Neutral and Happy occupies most the samples. If use all the samples to train the model, the model is likely to predict most sample into neutral and happy. The samples need to balance. After training the model with several iterations. a balancing sample training set should be prepared and used for continuing training. In SVM training, after training with all sample, we select the hard samples between the board line of each class pair and retrain the SVM model. This will improve the ability the recognition for fewer-samples classes.

## 5.4 Prediction results

Comparing with different classifiers, the feature (SIFT, LBP, geometry) are generated and form the feature vector. Without considering the pose factor, the expression recognition becomes harder and the accuracy is low. Our proposed network combines the hand-crafted features with the deep feature. The performance is better than only using deep net feature or hand-crafted features. The traditional classifiers also show great ability to deal with the problem, but not better than the CNN approach. For each pose, the comparison of the proposed network and the traditional classifiers are shown in Table 3.

|        | Classifiers |          |       | Proposed Network |                |
|--------|-------------|----------|-------|------------------|----------------|
|        | SVM         | Random F | ANN   |                  |                |
| Pose 1 | 77.3%       | 73.2%    | 73.3% | 80.3%            | No head pose   |
| Pose 2 | 71.7%       | 75.2%    | 69.2% | 78.71%           | separation:    |
| Pose 3 | 68.7%       | 75.1%    | 72.5% | 79.5%            | 66.7%          |

Table 3. Expression recognition in different poses by different classifiers

Comparing with different single feature, as Table 2.

|        | SVM Classifier |           |          |                  | Proposed net          |                  |
|--------|----------------|-----------|----------|------------------|-----------------------|------------------|
|        | SIFT           | Grid TPLBP| Geometry | Combined feature | No handcraft feature  | Combined feature |
| Pose 1 | 72.1%          | 74.5%     | 75.5%    | 75.8%            | 70.3%                 | 80.3%            |
| Pose 2 | 71.2%          | 72.6%     | 78.4%    | 78.4%            | 62.1%                 | 78.7%            |
| Pose 3 | 75.4%          | 79.9%     | 75.1%    | 79.9%            | 74.2%                 | 79.5%            |

Table 4. Expression recognition in different features

We list the confusion matrixes of the prediction.

Pose class 1, average accuracy: 80.3%. Confusion matrix as Table 5.

|          | Neutral | Happy | Sad | Fear | Angry | Surprise | Disgust |
|----------|---------|-------|-----|------|-------|----------|---------|
| Neutral  | 2397    | 261   | 10  | 17   | 19    | 26       | 6       |
| Happy    | 476     | 2681  | 7   | 2    | 11    | 9        | 1       |
| Sad      | 101     | 13    | 5   | 5    | 1     | 14       | 0       |
| Fear     | 94      | 18    | 1   | 25   | 1     | 2        | 0       |
| Angry    | 135     | 18    | 0   | 0    | 24    | 2        | 0       |
| Surprise | 82      | 14    | 2   | 0    | 2     | 61       | 1       |
| Disgust  | 126     | 32    | 6   | 1    | 4     | 8        | 9       |

Table 5. Confusion matrix of pose class 1 prediction

Pose class 2, average accuracy: 78.71%. Confusion matrix as Table 6

|  | Neutral | Happy | Sad | Fear | Angry | Surprise | Disgust |
|---|---|---|---|---|---|---|---|
| Neutral | 1997 | 205 | 50 | 28 | 35 | 41 | 5 |
| Happy | 459 | 1936 | 9 | 5 | 21 | 23 | 11 |
| Sad | 91 | 2 | 20 | 4 | 1 | 12 | 2 |
| Fear | 79 | 4 | 11 | 10 | 6 | 2 | 1 |
| Angry | 141 | 15 | 5 | 2 | 30 | 0 | 0 |
| Surprise | 151 | 21 | 9 | 8 | 6 | 48 | 5 |
| Disgust | 85 | 17 | 13 | 9 | 2 | 2 | 9 |

Table 6. Confusion matrix of pose class 1 prediction

Pose class 3, average accuracy: 79.5%. Confusion matrix as Table 7

|  | Neutral | Happy | Sad | Fear | Angry | Surprise | Disgust |
|---|---|---|---|---|---|---|---|
| Neutral | 1450 | 255 | 30 | 25 | 17 | 39 | 15 |
| Happy | 198 | 1869 | 12 | 14 | 7 | 26 | 6 |
| Sad | 68 | 10 | 24 | 7 | 3 | 3 | 3 |
| Fear | 27 | 18 | 0 | 36 | 0 | 14 | 0 |
| Angry | 103 | 7 | 5 | 3 | 39 | 0 | 0 |
| Surprise | 95 | 5 | 4 | 12 | 2 | 69 | 2 |
| Disgust | 471 | 16 | 14 | 3 | 1 | 5 | 8 |

Table 7. Confusion matrix of pose class 1 prediction

## 6. CONCLUSION

In our work, we present the method using convolutional neural networks to recognize facial expressions in extreme poses based on in-the-wild images. We combined SIFT, TPLBP and geometric feature to the net. From the result, we can see that the combined features improve the overall performance. As comparisons, we built multiclass SVM classifiers, random forest and artificial neural network, to do the same classification problem. Different classifiers have different strengths and drawbacks and the performance also varies. The aim of our work is to explorer the possible methods that can be implemented into strained facial expression recognition problems to apply on problems with in-the-wild extreme pose faces.

In our work, we separate the photos into only 5 pose classes without considering the head synthesized generating process of the data set, while the data set consists the head rotated photos with increased 10-degree yaw. That means we can get more accurate pose yaw degree and separate the photos to more pose classes. The benefit of this is that we can take a further step to minimize the effect of head pose for our trained classification models. One previous step to do this is to estimate the pose of the original photo accurately.

We only consider several left or right head poses, but in fact the head pose has many more poses like turn up or down and the synthesized pose of left-right and up-down [36]. Considering those effects, there would be more pose classes that need to define. And for each pose class, the classifier models should be trained. Under this situation, it is more difficult to just use PCA projection to separate the pose class, because the most remarkable difference between landmarks under that situation has not been the angle of left-right head pose. Pose estimation should be estimated more accurately according to the geometric features, relating to the landmark information [36][37]. And currently this lacks datasets.

About the expression labeling, we labeled the photos by three people and voted for the final labels. But the three people are not expression experts and even in this voting way, there are still much mislabeling happening, especially when for one face image, different people had totally different opinions. The way to solve the problem is to find more experts to label the dataset. Besides of this, the data set is much unbalanced in the number of different expressions. As talked, the majority of the data set samples are neutral or happy samples and there are much fewer samples for other expressions, which will cause the unbalanced dataset problems when training the classifier models.

The data set in our work is 300W-LP. It is the 3D generating series of images from one original photos. That is not real original extreme pose photos. During the training for the features extracted from these photos, we may think that there could be some repeated data existing in some certain feature dimensions. This is bad for the generalization of the training. And for the wrong labeling photos, the repeated samples will perform as a very bad actor affecting the final trained models. So further work can be to eliminate some repeated photos within one pose class. Ideally, it is good to have only one or two samples generated from the same photos existing in one pose class. Or we can also think that it is possible to propose another data set for facial expression recognition in-the-wild in extreme pose.

## 7. ACKNOWLEDGMENT

The author acknowledges the financial support from the International Doctoral Innovation Centre, Ningbo Education Bureau, Ningbo Science and Technology Bureau, and the University of Nottingham. This work was also supported by the UK Engineering and Physical Sciences Research Council [grant number EP/L015463/1].